\documentclass[sigconf]{acmart}
\def\BibTeX{{\rm B\kern-.05em{\sc i\kern-.025em b}\kern-.08emT\kern-.1667em\lower.7ex\hbox{E}\kern-.125emX}}

\copyrightyear{2020}
\acmYear{2020}
\setcopyright{acmcopyright}
\acmConference[CIKM '20]{Proceedings of the 29th ACM International Conference on Information and Knowledge Management}{October 19--23, 2020}{Virtual Event, Ireland}
\acmBooktitle{Proceedings of the 29th ACM International Conference on Information and Knowledge Management (CIKM '20), October 19--23, 2020, Virtual Event, Ireland}
\acmPrice{15.00}
\acmDOI{10.1145/3340531.3412701}
\acmISBN{978-1-4503-6859-9/20/10}

\usepackage{xcolor}
\usepackage[utf8]{inputenc}
\usepackage{amsmath}
\usepackage{multirow}
\usepackage{booktabs}
\usepackage[ruled, lined, linesnumbered, commentsnumbered, longend]{algorithm2e}
\usepackage{xcolor}

\usepackage[noend]{algpseudocode}
\def\algbackskip{\hskip-\ALG@thistlm}
\makeatother

\begin{document}
\fancyhead{}

\sloppy
\SetCommentSty{mycommfont}
%
\title{Efficient Neural Query Auto Completion}

\author{Sida Wang, Weiwei Guo, Huiji Gao, Bo Long}
\affiliation{
  \institution{LinkedIn, Mountain View, California}
}
\email{{sidwang, wguo, hgao, blong}@linkedin.com}

\renewcommand{\shortauthors}{Wang et al.}

%
\begin{abstract}
Query Auto Completion (QAC), as the starting point of information retrieval tasks, is critical to user experience. Generally it has two steps: generating completed query candidates according to query prefixes, and ranking them based on extracted features. Three major challenges are observed for a query auto completion system: (1) QAC has a strict online latency requirement. For each keystroke, results must be returned within tens of milliseconds, which poses a significant challenge in designing sophisticated language models for it. (2) For unseen queries, generated candidates are of poor quality as contextual information is not fully utilized. (3) Traditional QAC systems heavily rely on handcrafted features such as the query candidate frequency in search logs, lacking sufficient semantic understanding of the candidate.

In this paper, we propose an efficient neural QAC system with effective context modeling to overcome these challenges. On the candidate generation side, this system uses as much information as possible in unseen prefixes to generate relevant candidates, increasing the recall by a large margin. On the candidate ranking side, an unnormalized language model is proposed, which effectively captures deep semantics of queries. This approach presents better ranking performance over state-of-the-art neural ranking methods and reduces $\sim$95\% latency compared to neural language modeling methods. The empirical results on public datasets show that our model achieves a good balance between accuracy and efficiency. This system is served in LinkedIn job search with significant product impact observed.
\end{abstract}

%

%
\maketitle

\section{Introduction}
Query auto completion (QAC) \cite{Cai:16} is the standard component of search engines in industry. Given a prefix, the system returns a ranked list of completed queries that match users' intents.  Query auto completion enhances user experience in two ways: (1) it saves user keystrokes and returns search results in less time; (2) it guides users to type "better" queries that are more likely to lead to desirable search results.  For example, given a prefix "soft", "software engineer" is a better query than "software programmer", since the former is a more commonly used job title.

A typical QAC system takes a generate-then-rank two-step framework. The candidate generation component returns the most frequent queries that match the prefix, by memorizing the mapping from prefixes to queries based on search logs \cite{Bar-Yossef:11}. The candidate ranking component extracts features from candidates and uses them to produce the final ranking order. Both components do not involve intense computation so the whole process can be finished within several milliseconds, in order to meet online latency requirements.

However, this traditional approach does not fully exploit the context in the query prefix. For example, in the generation phase, for unseen prefixes, only the last word of prefix is used to generate candidates \cite{Mitra:15}; in the ranking phase, the most effective feature is the query frequency collected from search log, which lacks deep semantics understanding.

To enable semantic text understanding, neural networks based methods are applied to QAC. Early works \cite{Mitra:15} focus on the ranking stage: Convolutional Neural Networks (CNN) \cite{Shen:14} are adopted to measure the semantic coherence between the query prefix (user input) and suggested suffixes in the ranking phase. Recently, an end-to-end approach for both generation and ranking \cite{Park:17} is proposed: a neural language model is trained to measure the probability of a sequence of tokens. During decoding, candidate generation and ranking are performed for multiple iterations during beam search \cite{Lowerre1976TheHS}. While neural language models show better sequence modeling power over CNN, they could take up to 1 second \cite{Wang2018}, making productionization infeasible.

In this work, our goal is to build an QAC system with more effective query context utilization for real world search systems, while meeting the industry latency standard. We make improvements in the two-stage generate-then-rank framework: (1) In candidate generation, we extend a previous work \cite{Mitra:15} by incorporating more information from unseen prefixes to generate meaningful candidates. (2) In candidate ranking, we adopt neural language modeling, a more natural approach to model the coherence between a word and its previous sequence. To overcome the latency challenge, we optimize the latency by approximating computation of word probability with a much more efficient structure, reducing 95\% latency ($\sim$55ms to $\sim$3ms). Offline experiments on public datasets show significant improvement in terms of relevance and latency. We also train our model on the LinkedIn job search dataset and deploy it in production with CPU serving.


In summary, the contribution of this paper is
\begin{itemize}
  \item We developed an effective candidate generation method that maximizes recall of clicked queries through better context utilization.
  \item We developed an optimized neural language model for candidate ranking, which has similar sequence modeling power as general neural language models, yet with significantly lower serving latency.
  \item We deploy this efficient QAC system into commercial search engines, and observe significant product impact.
\end{itemize}

\section{Related Work}
In this session, we first introduce traditional methods for QAC, then discuss how neural networks are applied for QAC, followed by details of neural language models.

\subsection{Traditional Approaches for QAC}
Most of the previous works adopt a two-step framework -- candidate generation and candidate ranking \cite{Cai:16}. The former aims to increase recall, and the latter focuses on increasing precision. Candidate generation components return a list of completed queries for a given prefix. Most works are based on prefix-to-query statistics calculated from search logs.  For example, Most Popular Completion (MPC) \cite{Bar-Yossef:11} directly looks up the top N most frequent historical queries that start with the entire prefix. Mitra et al. \cite{Mitra:15} extend this method and generate candidates for rare or unseen prefixes by exploiting information from the last word in the prefix.  Other works investigate the problem of generating candidates directly from documents when search logs are not available \cite{bhatia2011query,maxwell:2017}.  In the candidate generation phase, we further extend \cite{Mitra:15} to use more context to generate better candidates.

After candidates are generated, candidate ranking components compute a relevance score for each candidate. Different sources of information have been exploited to improve the ranking, including context or session information \cite{Bar-Yossef:11,Jiang:14}, time/popularity-sensitivity \cite{Shokouhi:12,Whiting:14,Cai:14}, personalization \cite{Shokouhi:13,Cai:14}, user behaviors \cite{Li:14,Mitra:14,Hofmann:14,Zhang:15}, and click through logs \cite{Li:17}.  Our paper does not use any of these additional information and focuses on the effectiveness and efficiency of QAC given general query logs. Therefore, our work is orthogonal to methods using additional information and can benefit these methods as well.

\subsection{Deep Learning Approach for QAC}
\label{section:related-work-dl}
In order to tackle insufficient text understanding in traditional approaches, deep learning approaches are adopted in recent years.

One line of work is applying neural networks to extract semantic embeddings from queries and perform ranking. In \cite{Mitra:15}, Convolutional Latent Semantic Model (CLSM) \cite{Shen:14} is used for QAC ranking. This model applies Convolutional Neural Networks (CNN) \cite{LeCun:95} to extract an embedding for query prefix and suffix strings.  Then the cosine similarity score between the prefix and suffix embedding determines the ranking.

Another line of work is applying neural language models for both generation and ranking \cite{Park:17, Wang2018}. In such work, a Long Short Term Memory (LSTM) \cite{Hochreiter:97} based neural language model \cite{mikolov2010} is trained on complete queries. After that beam search \cite{Lowerre1976TheHS} is used for decoding. In beam search, there are many iterations of generation and ranking, which yield impressive relevance performance with a large computation overhead.  There are several advantages of neural language modeling. One advantage of the neural language modeling architecture is that additional features can be seamlessly incorporated.  For example, personalization can be modeled by incorporating user ID embeddings \cite{Jaech:18,Fiorini:18,Jiang:18} in the network. Time aware \cite{Fiorini:18} and spelling errors aware \cite{Wang2018} models are also developed under this framework. Another advantage is that language modeling is more effective at sequence coherence modeling, supported by its probabilistic interpretation $P(query)=\prod_i{P(w_i|w_0...w_{i-1})}$.

\subsection{Neural Language Modeling}
Neural language models measure the probability of a text sequence. Bengio et al.\ \cite{Bengio:03} propose a neural language model, where the probability of the next word is computed based on the embeddings of previous several words.  Mikolov et al.\ \cite{mikolov2010} use Recurrent Neural Networks (RNN) \cite{Williams:89}  to summarize a sequence of any length into a hidden state of RNN, which generalizes the context better. Following these works are word level \cite{sundermeyer2012} and character level \cite{Kim:16} neural language models.

However, these neural language modeling approaches are time consuming in both training and decoding stages -- in the computation of word probability, these methods compute a costly normalization term which requires iterating over all words in the vocabulary. In order to resolve this issue, unnormalized language models \cite{gutmann2010,Devlin:14} are proposed, targeting latency reduction. The idea is to approximate the normalization over the whole vocabulary in word probability computation. Unnormalized language models have been applied on machine translation \cite{Devlin:14}, speech recognition \cite{chen2015, sethy2015} and word embedding pretraining \cite{mnih2013}.  We propose an efficient unnormalized language model approach for QAC ranking, and deploy it into LinkedIn's search engines.

\section{An Efficient Neural Query Auto Completion System}
We propose an efficient neural query auto completion system that consists of two phases: candidate generation and candidate ranking. In candidate generation, we aim to increase recall of candidates with more context utilization. This step is finished within 1 millisecond. In candidate ranking, we design an efficient unnormalized neural language model that effectively models the query sequence. Therefore, the following content focuses on (1) how to exploit more contexts for both generation and ranking, as well as (2) how to minimize the computation cost in neural language model based ranking.

\subsection{Candidate Generation}
\label{sec:cg}

\begin{table}
  \caption{Candidate generation for the query prefix ``cheapest flights from seattle to".  The grayed-out words are not used in the candidate generation process, and the italicized words are the matched suffixes. The example shows that utilization of more words leads to stronger relatedness between suffixes and prefixes. \textit{E.g.}, hinted by "flights from seattle to", more relevant suggestions like "... to sfo" are suggested. In contrast, given only the last word "to", suggestions tend to have lower quality like ``airport".}
  \label{tab:cg-example}
  \begin{tabular}{l}
    \toprule
    cheapest flights from seattle to\\
    \midrule
    \textcolor{gray}{cheapest} {\it flights from seattle to sfo}\\
    \textcolor{gray}{cheapest flights from} {\it seattle to vancouver}\\
    \textcolor{gray}{cheapest flights from seattle} {\it to airport}\\
    \textcolor{gray}{cheapest flights from seattle} {\it to study}\\
  \bottomrule
\end{tabular}
\end{table}

The candidate generation component returns a list of queries for a given prefix. The first step is to collect {\it background} data within a certain time range. The background data contain queries and their corresponding frequency computed from search logs within the time range. The second step is to build a prefix-to-query mapping based on the background data. Given a prefix, candidates are generated by looking up the mapping and choosing the top N most frequent queries.  The whole process can be optimized in microseconds with the Apache Finite State Transducer (FST) library.\footnote{\url{https://lucene.apache.org/core/7_3_1/core/org/apache/lucene/util/fst/FST.html}}

\begin{figure*}[h]
  \includegraphics[width=120mm]{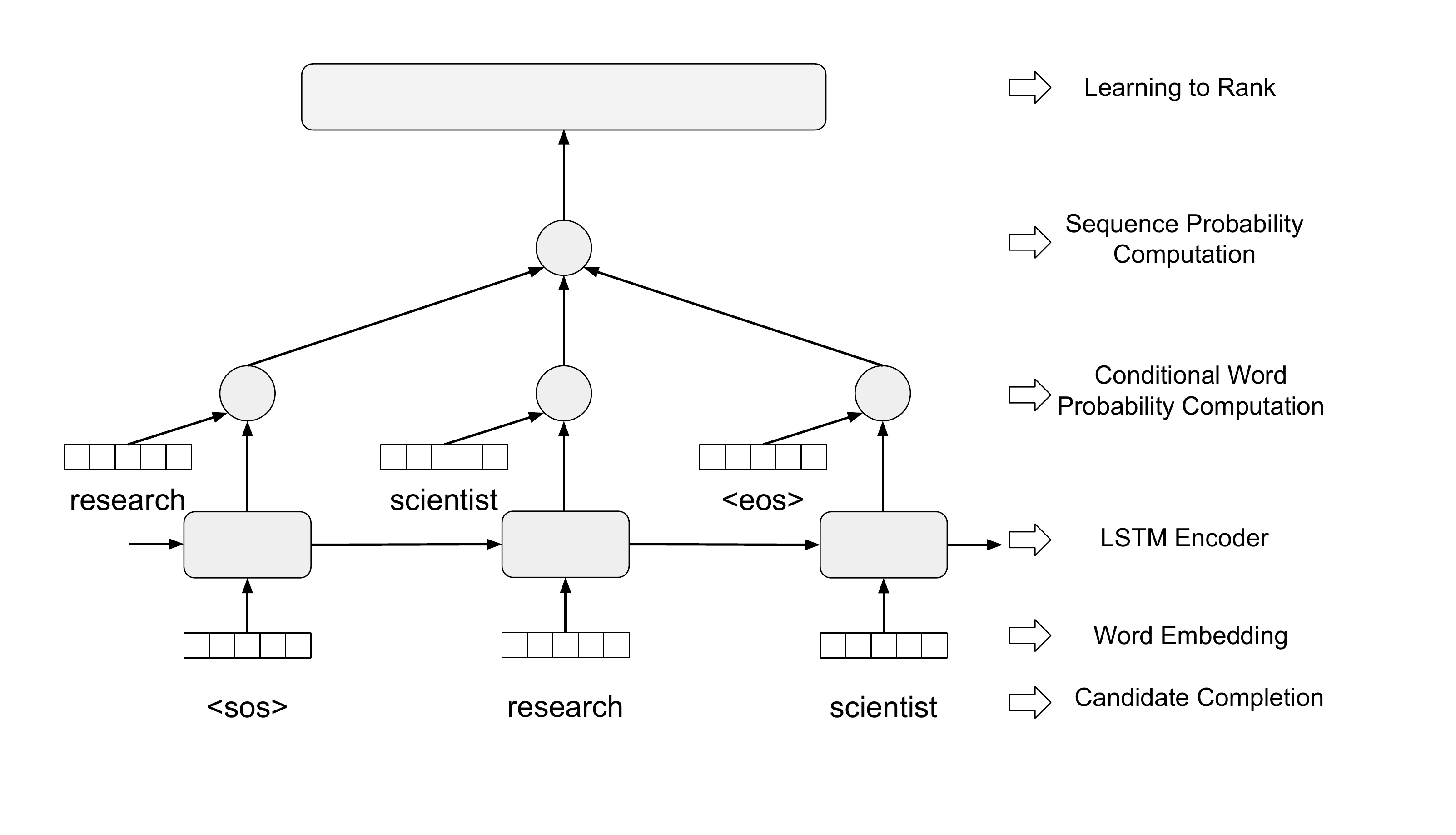}
  \caption{Our neural ranking model architecture.  On top of it is a Learning-To-Rank layer that takes in multiple candidate scores. The input query has a special token "<sos> research scientist"; the probability of "research scientist <eos>" is computed based on LSTM hidden states.}
  \label{fig:model}
\end{figure*}

One obvious issue is that the prefix may never be seen in the {\it background} data. Mitra and Craswell \cite{Mitra:15} overcome this issue by exploiting {\bf suffixes}. First, 100k most frequent suffixes are collected from the background queries.  Then, for an unseen user input such as  ``cheapest flights from seattle to", the algorithm extracts the last word ``to",\footnote{It could also be the last incomplete word.  For example, if the prefix is "cheapest flights from seattle t", then the last incomplete word is "t".} and uses it to match any suffixes that start with ``to". Therefore, they are able to find suffixes such as ``to dc", ``to bermuda", {\it etc.}, which can be appended to ``cheapest flights from seattle". We refer to this method as LastWordGeneration (LWG).

Note that two separate FSTs are built: one is built on background queries, and the other one is built on all suffix n-grams of the queries. We call these two FSTs \textbf{QueryFST} and \textbf{SuffixFST} respectively.

It's observed that the quality of retrieved suffixes is not high, when only the last word is used for matching. We believe using more contexts to match suffixes should yield more relevant results. Therefore, we extend this approach by greedily matching the longest last few words, instead of only the last one word. For the unseen prefix ``cheapest flights from seattle to", we first try to find suffixes that start with ``flights from seattle to", then ``from seattle to", ``seattle to", and finally "to".  An example is shown in Table \ref{tab:cg-example}.  By using more prefix tokens, it works effectively for recall improvement. This approach is referred to as \textbf{Maximum Context Generation (MCG)} and described in Algorithm \ref{alg:cg}.

\begin{algorithm}
    \SetKwFunction{len}{len}
    \SetKwInOut{KwIn}{Input}
    \SetKwInOut{KwOut}{Output}

    \KwIn{QueryFST \textit{$F_q$}, SuffixFST \textit{$F_s$}, query prefix \textit{p}}
    \KwOut{An ordered list of queries \textit{qList}}

    \tcc{1. Get suggestions from QueryFST}
    $qList = F_q(p)$
    
    \tcc{2. Add suggestions from SuffixFST}
    
    $rw = $ ""  \tcp*[f]{Keep record of removed words} \\
    \While{$p$ \text{!= ""}}
   {
        \tcp{Remove first word $w_0$ from current prefix}
        $w_0 = p.remove(0)$ \\
        $rw = rw + w_0$ \\
        \tcp{Prepend removed words to suggestions from SuffixFST}
   		$sList = [rw + s\ \ for\ \ s\ \ in\ \ F_s(p)]$ \\
   		$qList.addAll(sList)$
   }

    \KwRet{$qList$}
    \caption{Maximum Context Generation}
    \label{alg:cg}
\end{algorithm}

\subsection{Candidate Ranking}
In this section, we focus on how to rank the generated candidates with neural networks.  As shown in Figure \ref{fig:model}, our system consists of two major components: an "unnormalized" language modeling \cite{Devlin:14,sethy2015} layer and a pairwise learning-to-rank \cite{Burges:05} layer. The unnormalized language model layer computes a score for a query candidate efficiently. Then learning-to-rank objective functions are applied on the scores of the clicked and non-clicked query pairs. These two components are trained together in an end-to-end fashion.

To score a query, an intuitive way is neural language modeling with LSTM \cite{Park:17} that computes the sequence probability as the query scores, as previously discussed in Section \ref{section:related-work-dl}. The log probability of the query is computed as:
\begin{align}
\label{equation:prob}
\log{P(q)} &= \sum_i^{L}{\log{P(w_i|w_1,w_2,...,w_{i-1})}} \nonumber\\
     &= \sum_i^{L}{\log{P(w_i|h_{i-1})}} \nonumber\\
     &= \sum_i^{L}{\log{\frac{e^{v_i^{\top}h_{i-1}}}{\sum_j e^{v_j^{\top}h_{i-1}}}}} \nonumber\\
     &= \sum_i^{L}{\left(v_i^{\top}h_{i-1} - \log{\sum_j^N{e^{v_j^{\top}h_{i-1}}}}\right)}
\end{align}
where $h_i$ is the hidden state of an LSTM \cite{Hochreiter:97} cell for word $w_i$, $v_i$ is the embedding for word $w_i$, $L$ is the number of words in the query and $N$ is the vocabulary size.  In this case, the hidden state $h_{i-1}$ summarizes all information of the sequence $w_1,w_2,...,w_{i-1}$ before $w_i$.  It is worth noting that two special tokens <sos> and <eos> are used, as illustrated in Figure \ref{fig:model}.

However, this natural language modeling approach is inefficient. To compute the probability of a word $w_i$, we need to compute the normalization term $\log{\sum_j^{N}{e^{v_j^{\top}h_{i-1}}}}$. This term involves vector multiplication between the hidden state $h_{i-1}$ and each word in the vocabulary. Usually the size of vocabulary can be larger than 30k, which produces a large computation overhead. Therefore, to reduce latency, approximation needs to be applied on the normalization term, similar to "unnormalized" language modeling. Such approximation must satisfy several requirements:

\begin{itemize}
  \item Computational efficiency. The approximation should not require iterating every word over the vocabulary.
  
  \item Ranking effectiveness. Under the approximation, although the absolute value of query probability $P(q)$ will be affected, it shall preserve its relative ranking position w.r.t. other queries in most cases.
  
  \item Length penalty. Long queries should receive more penalty than short ones, as short queries are more frequently typed.\footnote{The same pattern (short queries are more frequent) is observed in both AOL and LinkedIn datasets.}
\end{itemize}

Therefore, we propose our design of unnormalized language modeling, where the normalization term is approximated by:

\begin{align}
    \log{\sum_j^N{e^{v_j^{\top}h_{i-1}}}} \approx b
\end{align}
where $b$ is a scalar parameter to learn. Accordingly, equation \ref{equation:prob} becomes:

\begin{align}
\label{equation:approx}
    \log{P(q)} \approx \sum_{i}^{L}{\left(v_i^{\top}h_{i-1}-b\right)}
\end{align}

This design satisfies the requirements: 1) For computational efficiency, since the approximation term is only a scalar value, latency is significantly reduced. 2) This design also keeps ranking effectiveness. 
More specifically, Equation \ref{equation:approx} measures the semantic coherence of words in the query by $\sum_{i}{v_i^{\top}h_{i-1}}$, hence it assigns a meaningful ranking score for the query.
3) It also imposes length penalty by using a penalty term $b*L$.

With this approximation, the latency can be reduced by more than 95\% in our experiments (Table \ref{tab:latency}).  Similar speedup is observed in other unnormalized language model methods \cite{gutmann2010,Devlin:14}.

\subsection{Comparison to State-of-the-Art Models}
\subsubsection{Comparison to CLSM Model}
CLSM is used for candidate ranking in the previous work \cite{Mitra:15}. This model applies Convolutional Neural Networks (CNN) \cite{LeCun:95} to learn an embedding for query prefix and suffix.  The coherence score is the cosine similarity score between the prefix and suffix embedding. However, CNN only focuses on extracting n-gram features that is useful for ranking; it ignores the sequence length, and does not explicitly model word coherence $P(w_i|w_0...W_{i-1})$. In our approach, we use neural language models that better model the coherence between words.  We train the neural language model with a learning-to-rank layer in an end-to-end approach.

\subsubsection{Comparison to an End-to-End Language Modeling Approach}
The end-to-end language modeling approaches \cite{Park:17, Wang2018} yield great performance, since generation and ranking are performed at the same time for many iterations (one iteration for one generated token). However, it is also time-consuming. For a character level language model, ranking is required in every character generation, compared to the first-generate-then-rank framework.

\subsubsection{Comparison to Existing Unnormalized Language Models} 
Different designs of unnormalized language models are proposed for machine translation \cite{Devlin:14,vaswani2013} and speech recognition \cite{sethy2015,chen2015} to reduce computation time.  In these applications, a valid probability is crucial for candidate generation. The first method is self normalization \cite{Devlin:14,sethy2015}, which adds a regularization term to the task specific loss, to make sure the sum of probability of all words is close to 1.
During decoding, the normalization term is dropped to accelerate speed, assuming it is equals to 1.  However, the disadvantage is that training is slow, since the normalization term needs to be computed in training to obtain the loss.

Another popular method is Noise-Contrastive Estimation \cite{gutmann2010,mnih2012fast,vaswani2013,chen2015}. For each word, $k$ noise words are added as negative examples. This method still requires a certain amount of the training time. Moreover, the normalization term is still computed at inference time and thus it does not meet the industrial latency requirement. In our case, our focus is the relative ranking of sentences (guided by the learning-to-rank loss) and both training and inference efficiency, rather than having an accurate estimation of the word probability. Therefore, we do not compare our model with Noise-Contrastive Estimation in this work.

\section{Experiment Setup}

\subsection{Dataset Preparation}
\label{sec:aol-data}
All experiments are conducted on the publicly available AOL dataset \cite{pass2006picture}. Similar results are observed on the LinkedIn job search dataset.

The data preprocessing follows the previous work \cite{Mitra:15}. We use the AOL data collected over three months (from March to June 2006) for experimentation. For preprocessing, we remove the empty query and keep only one copy of the adjacent identical queries. The dataset is split in the same way as in \cite{Mitra:15} -- data from March 1 to April 30, 2006 are used as background data, with the following two weeks as training data, and each of the following two weeks as validation and test data. This results in 13.88 million background queries, 3.19 million training queries, 1.37 million validation queries, and 1.40 million test queries. For robustness, we exclude queries with frequency < 3 in the background data.

Mapping from prefix to queries is constructed and stored in two FSTs, QueryFST and SuffixFST, as described in Section \ref{sec:cg}: QueryFST is built on the background data; SuffixFST is built on the most frequent 100k suffixes in the background data. No special treatment is applied on out-of-vocabulary words such as mapping them to unknown tokens. If a word cannot be found in FSTs, no suggestion will be made.

\subsection{Baseline Models in Candidate Generation}

In candidate generation, MostPopularCompletion (MPC) \cite{Bar-Yossef:11} and LastWordGeneration (LWG) \cite{Mitra:15} are used as baselines. These two methods are state-of-the-art methods that focus on usage of query prefixes and do not take in additional information such as personalization and time awareness.

\textbf{MostPopularCompletion (MPC):} Given a query prefix $p$, this method searches for the most frequent $k$ ($k$ is the number of candidates to be ranked in candidate ranking) queries starting with $p$ in QueryFST. QueryFST is built in the same way as in Section \ref{sec:aol-data}.

\textbf{LastWordGeneration (LWG):} Given a query prefix $p=w_1w_2...w_n$, this method first obtains historical queries starting with $p$ from QueryFST like MPC. After that, the last word $w_n$ (it could be an incomplete word) is extracted and the most frequent $k$ suffixes starting with $w_n$ are collected from SuffixFST. Extracted suffixes are prepended with $w_1w_2...w_{n-1}$ (words in prefix except $w_n$) to make up a suggestion. Finally candidates from QueryFST and SuffixFST are merged together.
QueryFST and SuffixFST are built in the same way as in Section \ref{sec:aol-data}.

\subsection{Baseline Models in Candidate Ranking} \label{sec:cr}
In candidate ranking, we compare our unnormalized language model to two categories of models, frequency based models and neural network models.
\subsubsection{Frequency based models}\label{sec:cr-freq} Frequency based models give higher ranks to more frequent candidates from MPC, LWG and MCG. For MPC, more frequent candidates are ranked higher. For LWG, candidates from the same FST are ranked by frequency. Candidates from QueryFST are ranked higher than those from SuffixFST. For MCG, the ranking is best described by Algorithm \ref{alg:cg}.

\subsubsection{Neural network models} Given results from candidate generation, neural network models generate a score for each candidate and rank the candidates by their scores. We compare our model to the state-of-the-art Convolutional Latent Semantic Model (CLSM) \cite{Shen:14} , and implement a simple LSTM model for comparison.

\textbf{CLSM}: Given a sequence and its corresponding embedding matrix, this model first extracts contextual features from n-grams using convolution filters and then extracts salient n-gram features using a max pooling layer. A semantic (dense) layer is then applied on the salient n-gram features to obtain a semantic vector for the sequence. Semantic vectors for both prefix and suggested suffix are extracted. Cosine similarity between these two vectors is computed and treated as the candidate score. In our experiments, all hyperparameters follow the same setting as in \cite{Mitra:15}.

\textbf{LSTMEmb}: The basic architecture of this model is an LSTM network. Given a word sequence, each word is fed into the LSTM cell in order. The final hidden state vector is used as the semantic representation of the sequence. Given this semantic vector, dot product is computed between the semantic representation and a learnable weight vector and used as the ranking score.

For LSTMEmb and our unnormalized language model, we choose an embedding size of 100 as a balance of performance and speed. Xavier initialization \cite{glorot2010understanding} is used for embeddings. The hidden state size is set to be the same as the embedding size. The AdamWeightDecay \cite{Loshchilov2019DecoupledWD} optimizer is used in training with a learning rate of $2*10^{-3}$ and a weight decay rate of $1*10^{-2}$. The vocabulary size is 30k. Word hashing is not applied in vocabulary generation. Pairwise ranking is used with logistic loss in the learning-to-rank layer. The scoring model and the learning-to-rank model are jointly trained. The maximum size of a candidate list is set to 10 for each user input.

\subsection{Evaluation}
Model comparison is conducted in terms of relevance and efficiency. In candidate generation, We use recall to measure the performance of candidate generation methods because this metric measures the probability that users' desired queries exist in the candidate list. More specifically, recall among top 10 results is measured.

In candidate ranking, we use mean reciprocal rank (MRR) to measure the relevance performance of each candidate ranking method. MRR is computed for the top 10 candidates. Similar to the previous work \cite{Mitra:15}, MRR is calculated for two groups: 1) seen prefixes, prefixes that have matches in QueryFST and 2) unseen prefixes, prefixes that cannot find matches from QueryFST and therefore completed queries only come from SuffixFST.

Latency is used as the measure of candidate generation efficiency. The time cost of ranking a candidate list with 10 candidates is measured for each model. This is the average time cost over 1000 tests. The average number of words in candidates is 3.20.

\section{Results on AOL}
\subsection{Candidate Generation}

\begin{table}[h]
  \caption{Performance of different candidate generation methods on AOL. For each method, candidates are generated in the same order as the ranking order described in Section \ref{sec:cr-freq}. Recall@10 is computed for all prefixes, seen prefixes and unseen prefixes separately. 
  $\dag$ indicates statistically significant improvements over LastWordGeneration through a paired t-test with p < 0.05.}
  \label{tab:cgen}
  \resizebox{0.45\textwidth}{!}{
  \begin{tabular}{lcccc}
    \toprule
    \multirow{2}{*}{\textbf{Candidate Generation Methods}} & \multicolumn{3}{c}{\textbf{Recall@10}}\\
    & All & Seen & Unseen\\
    \midrule
    MostPopularCompletion (MPC) & 0.2075 & 0.5091 & 0.0000 \\
    LastWordGeneration (LWG) & 0.3884 & 0.5207 & 0.2973 \\
    MaximumContextGeneration (MCG) & \textbf{0.3992$\dag$} & \textbf{0.5219$\dag$} & \textbf{0.3147$\dag$} \\
  \bottomrule
\end{tabular}
}
\end{table}

Table \ref{tab:cgen} compares the performance of different candidate generation methods, namely MPC, LWG and MCG. In consideration of efficiency, there's a limit on the number of candidates to be generated and ranked. In our experiment, this limit is set to be 10. For each method, candidates are generated in the same order as the ranking order described in Section \ref{sec:cr-freq}. Recall@10 is computed and its value is the same as the maximum MRR@10 that can be obtained in candidate ranking.

Our proposed candidate generation method MCG, has shown lift of Recall@10 over MPC and LWG both in total and in each prefix partition, with the major lift on unseen suffixes. In comparison with MPC, our methods provide a solution for rare and unseen suffixes, effectively leveraging context information of sequences following the first word of the prefix through the use of SuffixFST. In comparison with LWG, although technically MCG and LWG can produce the same candidate set, the candidate generation orders are different. Given the limit on the maximum number of candidates, this order has a large impact on QAC system performance. Our method prioritizes the generation of candidates that share more context with the user input. Therefore, MCG is able to exploit more context in user input to generate candidates with higher quality.

\subsection{Candidate Ranking}

\begin{table*}
  \caption{Performance of different candidate ranking models on AOL. Frequency based models are applied on each generation method. For each neural model, two settings are performed: (1)  use the network to rank all candidates; (2) only use the neural network to rank candidates generated by SuffixFST (noted by "model-variant"+Frequency). For methods involving neural networks, percentage lift is computed related to CLSM and CLSM + Frequency respectively. $\dag$ indicates statistically significant improvements over CLSM and $\ddag$ indicates statistically significant improvements over CLSM + Frequency through a paired t-test with p < 0.05.}
  \label{tab:rerank}
  \resizebox{0.65\textwidth}{!}{
  \begin{tabular}{lllcc}
      \toprule
    \multirow{2}{*}{\textbf{Generation}} & \multirow{2}{*}{\textbf{Ranking}} & \multicolumn{3}{c}{\textbf{MRR@10}}  \\
    & & All & Seen & Unseen \\
    \midrule
    MPC & Frequency & 0.1805 & 0.4431 & 0.0000 \\
    \midrule
    LWG & Frequency & 0.3147 & 0.4465 & 0.2241 \\
    \midrule
    MCG & Frequency & 0.3283	& 0.4469 & 0.2467 \\
    \cmidrule{2-5}
    & CLSM & 0.3270 & 0.4229& 0.2610 \\
    & LSTMEmbed &0.3278$\dag$ (+0.244\%)&	0.4224&	0.2628$\dag$  \\
    & UnnormalizedLM & 0.3328$\dag$ (+1.769\%) &	0.4293$\dag$&	0.2665$\dag$  \\
    & NormalizedLM & 0.3331$\dag$ (+1.865\%) & 0.4293$\dag$&	0.2669$\dag$  \\
    
    \cmidrule{2-5}
    & CLSM + Frequency & 0.3369 &	0.4472 &	0.2610 \\
    & LSTMEmbed +Frequency & 0.3379$\ddag$ (+0.297\%) &	0.4472	& 0.2628$\ddag$ \\
    & \textbf{UnnormalizedLM +Frequency} & \textbf{0.3402$\ddag$ (+0.980\%)} & \textbf{0.4473}	& \textbf{0.2665$\ddag$} \\
    & \textbf{NormalizedLM +Frequency} & \textbf{0.3404$\ddag$ (+1.039\%)} & \textbf{0.4473}	& \textbf{0.2669$\ddag$ }\\
  \bottomrule
\end{tabular}
}
\end{table*}

Table \ref{tab:rerank} shows the performance of frequency based models, neural models and hybrid models. Frequency based models and neural models are described in Section \ref{sec:cr}. Hybrid models are the combination of neural and frequency based models, denoted by NN+Frequency in Table \ref{tab:rerank}, where NN is a neural model. Among them, neural models only rank candidates from SuffixFST while keeping positions of candidates from QueryFST ranked by frequency.

As shown in the previous section, MCG exhibits the best candidate generation performance. Therefore, all neural ranking methods are performed on the candidates generated by MCG.

Consistent with results from the prior work \cite{Park:17}, neural models that rank all candidates cannot outperform frequency-based methods on seen user inputs. This shows neural networks' insufficiency in memorizing strong evidence. However, all neural networks exhibit lift on unseen user inputs, showing their power in evaluating the coherence of the query and the semantic relation between suggested suffixes and words not used by SuffixFST (i.e., words "cheapest flights from" and the suffix "seattle to vancouver" in Table \ref{tab:cg-example}).

Based on this observation, we conduct experiments for hybrid models. In these models, we keep the ranking of candidates from QueryFST given by frequency-based methods and apply neural networks only on candidates from SuffixFST. Results show that such combination achieves the best performance, with lift not only on unseen user inputs but also on seen user inputs. Note that the lift on seen user inputs comes from the fact that for seen user inputs, there are also results from SuffixFST when QueryFST provides less than 10 results.

Among results from neural models except normalized neural language models, our neural unnormalized language model performs the best both with and without the combination with frequency-based methods. This gain comes mainly from the design of language modeling, a more effective context modeling architecture.

\begin{figure*}[h]
  \centering
  \includegraphics[width=140mm]{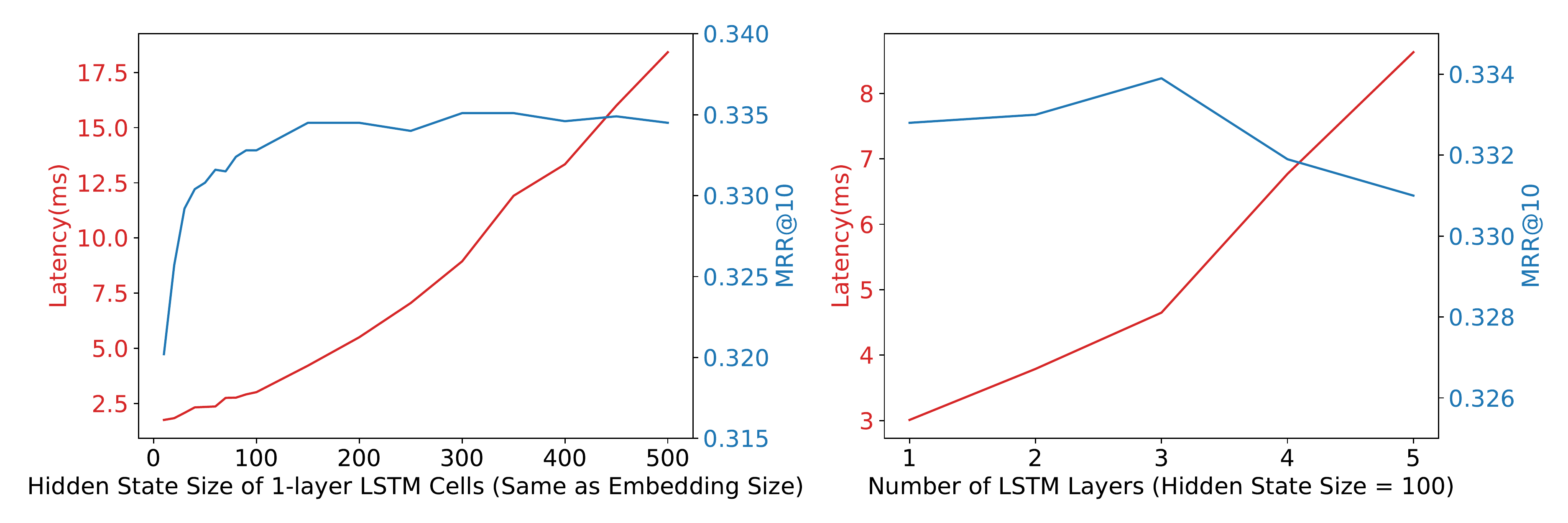}
  \caption{\small Latency \& MRR regarding LSTM hidden size (left) and latency \& MRR regarding LSTM layer number (right).}
  \label{fig:emb_lat}
\end{figure*}

\subsection{Latency}
The latency of MCG and neural ranking methods is shown in Table \ref{tab:latency}. The time cost of MCG is almost negligible because humans generally cannot notice latency in sub-millisecond level in auto completion. This shows the efficiency of MCG in capturing user context. Since the order of generated completions from MCG follows the order of frequency (high to low), the frequency based method with MCG generation has the same latency as MCG. Therefore, the major time cost of our two-step QAC system comes from neural ranking.

Compared with CLSM, our proposed unnormalized language model takes slightly more time, but it is still at the same scale. The extra time cost comes from the sequential dependency of LSTM cell outputs -- output of each LSTM cell depends on the state of the previous cell. This is not an issue for CNN as the convolution can be done in parallel. Under the current setting of candidate number and length, our model only takes 3 ms to rank candidates, indicating the unnormalized model is efficient for industrial applications.

To get a sense of the speed of normalized neural language modeling, we also implement a normalized neural language model. This model computes the exact conditional probabilities of word occurrences without any approximation. Results show that a normalized neural language model takes ~17 more times than an unnormalized language model. An average latency cost of 53.32ms for model inference is hardly acceptable for an auto completion system. For a fully generation-based neural language model with beam search, the latency will be further increased by a factor of beam search size. 

\section{Results on LinkedIn Production System}
We apply the model on LinkedIn job search dataset, and conduct A/B tests with 50\% traffic for two weeks. In the online system, users will receive QAC suggestions while typing. A list of job postings will be retrieved after user queries are submitted. Users can then view the postings and apply for jobs.

Figure \ref{fig:emb_lat} investigates the impact of the embedding size/hidden size and the layer number of LSTM on model inference speed and relevance performance. Even with the embedding size and the hidden vector size up to 500, the latency of our model is still lower than half of the latency of a normalized language model. Therefore, compared to normalized language models, under the same latency requirement, our unnormalized language modeling design can support more advanced models like Transformers \cite{vaswani2017attention} which have a better context understanding capability.

\begin{table}
  \caption{\small Latency of different models. The average time cost of ranking a candidate list with 10 candidates is measured for each model. The average number of words in candidates is 3.20. The hidden vector size and embedding size of LM is 100 and the LSTM layer number is 1. CLSM parameters are the same as in \cite{Mitra:15}. This test is conducted on a Intel(R) Xeon(R) CPU E5-2620 v3 @ 2.40GHz machine with 6 cores and 64-GB memory.}
  \label{tab:latency}
  \resizebox{0.32\textwidth}{!}{
  \begin{tabular}{lc}
    \toprule
    \textbf{Methods} & \textbf{Latency} \\
    \midrule
    MaximumContextGeneration & 0.18ms \\
    \midrule
    CLSM & 2.15ms \\
    Unnormalized LM & 3.01ms \\
    Normalized LM & 53.32ms \\
  \bottomrule
\end{tabular}
}
\end{table}

\begin{table}
  \caption{\small Job Search Online Results. All 3 metrics are statistically significant (p < 0.05).}
  \label{tab:js-ol-result}
  \resizebox{0.20\textwidth}{!}{
  \begin{tabular}{lc}
    \toprule
    \textbf{Metric} & \textbf{Lift}\\
    \midrule
    QAC CTR@5 & +0.68\% \\
    \midrule
    Job Views & +0.43\% \\
    Job Apply Clicks & +1.45\% \\
  \bottomrule
\end{tabular}
}
\end{table}

\vspace{1mm}
\noindent\textbf{Dataset Preparation:} We use one-month click through data from the LinkedIn job search engine to conduct experiments.  The data splitting (background, training, validation and test data) and FST construction are done in the same way as that in AOL. 

\vspace{1mm}
\noindent\textbf{Baseline System: } The baseline system follows a two-step ranking framework, with MCG as the candidate generation method and XGBoost \cite{chen2016xgboost} as the candidate ranking method. Multiple features are included in the XGBoost ranking model, such as frequency of suggested queries from background query logs, a Kneser–Ney smoothing language model score of the suggested queries, etc.

\vspace{1mm}
\noindent\textbf{Metrics: }
We measure the performance in two aspects: (1) the impact on query auto completion component, by \textbf{QAC CTR@5} (the click through rate of the top 5 suggested completions); (2) the impact on the overall job search results, by \textbf{Job Views} (number of jobs that users view through job search) and \textbf{Job Apply Clicks} (number of jobs that users apply through job search).






\vspace{1mm}
\noindent\textbf{Online Results: } We performA/B tests between model \textit{UnnormalizedLM +Frequency} and the baseline system as shown in Table \ref{tab:js-ol-result}. Since the baseline system uses MCG in candidate generation as well, the focus of online experiments is on comparing the performance of a hand-crafted feature based model to an unnormalized neural language model. The QAC CTR@5 metric lift indicates that the quality of query auto completion is improved. The Job Views/Job Apply Clicks metric lifts show that more relevant job postings are retrieved because more meaningful queries are issued. The model is ramped to 100\% traffic at LinkedIn's English job search.

\section{Conclusions and Future Work}

In this paper, we propose an efficient neural QAC system that captures contextual information in both candidate generation and ranking given general logs. Our method is orthogonal to models that use personalized information such as user search history.

On the candidate generation side, more context words in query prefixes are utilized, resulting in more relevant candidates, especially for unseen prefixes. On the candidate ranking side, an unnormalized language model is proposed, which enables real-time deep semantic understanding of queries. 

Besides its success in offline experiments, this system has been applied on the LinkedIn platform with great success. This technology not only saves user effort by suggesting queries related to users' intent, but also helps users better reach their goals by providing queries that are more likely to retrieve desirable results.

In the future, we would like to explore acceleration techniques like \cite{Wang2018}. This direction is of great importance because it enables more advanced NLP techniques in an industrial QAC system. E.g., normalized neural language generation, an end-to-end QAC system with high recall \cite{Park:17}, can be productionized. More advanced semantics encoders such as BERT \cite{devlin2018bert} can be used as well.

\bibliographystyle{ACM-Reference-Format}
\bibliography{nac}


\begin{thebibliography}{43}


\ifx \showCODEN    \undefined \def \showCODEN     #1{\unskip}     \fi
\ifx \showDOI      \undefined \def \showDOI       #1{#1}\fi
\ifx \showISBNx    \undefined \def \showISBNx     #1{\unskip}     \fi
\ifx \showISBNxiii \undefined \def \showISBNxiii  #1{\unskip}     \fi
\ifx \showISSN     \undefined \def \showISSN      #1{\unskip}     \fi
\ifx \showLCCN     \undefined \def \showLCCN      #1{\unskip}     \fi
\ifx \shownote     \undefined \def \shownote      #1{#1}          \fi
\ifx \showarticletitle \undefined \def \showarticletitle #1{#1}   \fi
\ifx \showURL      \undefined \def \showURL       {\relax}        \fi
\providecommand\bibfield[2]{#2}
\providecommand\bibinfo[2]{#2}
\providecommand\natexlab[1]{#1}
\providecommand\showeprint[2][]{arXiv:#2}

\bibitem[\protect\citeauthoryear{Bar-Yossef and Kraus}{Bar-Yossef and
  Kraus}{2011}]%
        {Bar-Yossef:11}
\bibfield{author}{\bibinfo{person}{Ziv Bar-Yossef} {and} \bibinfo{person}{Naama
  Kraus}.} \bibinfo{year}{2011}\natexlab{}.
\newblock \showarticletitle{Context-sensitive query auto-completion}. In
  \bibinfo{booktitle}{\emph{WWW}}.
\newblock


\bibitem[\protect\citeauthoryear{Bengio, Ducharme, Vincent, and Jauvin}{Bengio
  et~al\mbox{.}}{2003}]%
        {Bengio:03}
\bibfield{author}{\bibinfo{person}{Yoshua Bengio}, \bibinfo{person}{Réjean
  Ducharme}, \bibinfo{person}{Pascal Vincent}, {and} \bibinfo{person}{Christian
  Jauvin}.} \bibinfo{year}{2003}\natexlab{}.
\newblock \showarticletitle{A neural probabilistic language model}.
\newblock \bibinfo{journal}{\emph{JMLR}} (\bibinfo{year}{2003}).
\newblock


\bibitem[\protect\citeauthoryear{Bhatia, Majumdar, and Mitra}{Bhatia
  et~al\mbox{.}}{2011}]%
        {bhatia2011query}
\bibfield{author}{\bibinfo{person}{Sumit Bhatia}, \bibinfo{person}{Debapriyo
  Majumdar}, {and} \bibinfo{person}{Prasenjit Mitra}.}
  \bibinfo{year}{2011}\natexlab{}.
\newblock \showarticletitle{Query suggestions in the absence of query logs}. In
  \bibinfo{booktitle}{\emph{SIGIR}}.
\newblock


\bibitem[\protect\citeauthoryear{Burges, Shaked, Renshaw, Lazier, Deeds,
  Hamilton, and Hullender}{Burges et~al\mbox{.}}{2005}]%
        {Burges:05}
\bibfield{author}{\bibinfo{person}{Chris Burges}, \bibinfo{person}{Tal Shaked},
  \bibinfo{person}{Erin Renshaw}, \bibinfo{person}{Ari Lazier},
  \bibinfo{person}{Matt Deeds}, \bibinfo{person}{Nicole Hamilton}, {and}
  \bibinfo{person}{Greg Hullender}.} \bibinfo{year}{2005}\natexlab{}.
\newblock \showarticletitle{Learning to rank using gradient descent}.
\newblock \bibinfo{journal}{\emph{ICML}}.
\newblock


\bibitem[\protect\citeauthoryear{Cai and De~Rijke}{Cai and De~Rijke}{2016}]%
        {Cai:16}
\bibfield{author}{\bibinfo{person}{Fei Cai} {and} \bibinfo{person}{Maarten
  De~Rijke}.} \bibinfo{year}{2016}\natexlab{}.
\newblock \showarticletitle{A survey of query auto completion in information
  retrieval}.
\newblock \bibinfo{journal}{\emph{Foundations and Trends{\textregistered} in
  Information Retrieval}} (\bibinfo{year}{2016}).
\newblock


\bibitem[\protect\citeauthoryear{Cai, Liang, and De~Rijke}{Cai
  et~al\mbox{.}}{2014}]%
        {Cai:14}
\bibfield{author}{\bibinfo{person}{Fei Cai}, \bibinfo{person}{Shangsong Liang},
  {and} \bibinfo{person}{Maarten De~Rijke}.} \bibinfo{year}{2014}\natexlab{}.
\newblock \showarticletitle{Time-sensitive personalized query auto-completion}.
  In \bibinfo{booktitle}{\emph{CIKM}}.
\newblock


\bibitem[\protect\citeauthoryear{Chen and Guestrin}{Chen and Guestrin}{2016}]%
        {chen2016xgboost}
\bibfield{author}{\bibinfo{person}{Tianqi Chen} {and} \bibinfo{person}{Carlos
  Guestrin}.} \bibinfo{year}{2016}\natexlab{}.
\newblock \showarticletitle{Xgboost: A scalable tree boosting system}. In
  \bibinfo{booktitle}{\emph{Proceedings of the 22nd acm sigkdd international
  conference on knowledge discovery and data mining}}. ACM.
\newblock


\bibitem[\protect\citeauthoryear{Chen, Liu, Gales, and Woodland}{Chen
  et~al\mbox{.}}{2015}]%
        {chen2015}
\bibfield{author}{\bibinfo{person}{Xie Chen}, \bibinfo{person}{Xunying Liu},
  \bibinfo{person}{Mark~JF Gales}, {and} \bibinfo{person}{Philip~C Woodland}.}
  \bibinfo{year}{2015}\natexlab{}.
\newblock \showarticletitle{Recurrent neural network language model training
  with noise contrastive estimation for speech recognition}. In
  \bibinfo{booktitle}{\emph{ICASSP}}. IEEE.
\newblock


\bibitem[\protect\citeauthoryear{Devlin, Chang, Lee, and Toutanova}{Devlin
  et~al\mbox{.}}{2018}]%
        {devlin2018bert}
\bibfield{author}{\bibinfo{person}{Jacob Devlin}, \bibinfo{person}{Ming-Wei
  Chang}, \bibinfo{person}{Kenton Lee}, {and} \bibinfo{person}{Kristina
  Toutanova}.} \bibinfo{year}{2018}\natexlab{}.
\newblock \showarticletitle{Bert: Pre-training of deep bidirectional
  transformers for language understanding}.
\newblock \bibinfo{journal}{\emph{arXiv preprint arXiv:1810.04805}}
  (\bibinfo{year}{2018}).
\newblock


\bibitem[\protect\citeauthoryear{Devlin, Zbib, Huang, Lamar, Schwartz, and
  Makhoul}{Devlin et~al\mbox{.}}{2014}]%
        {Devlin:14}
\bibfield{author}{\bibinfo{person}{Jacob Devlin}, \bibinfo{person}{Rabih Zbib},
  \bibinfo{person}{Zhongqiang Huang}, \bibinfo{person}{Thomas Lamar},
  \bibinfo{person}{Richard Schwartz}, {and} \bibinfo{person}{John Makhoul}.}
  \bibinfo{year}{2014}\natexlab{}.
\newblock \showarticletitle{Fast and Robust Neural Network Joint Models for
  Statistical Machine Translation}.
\newblock \bibinfo{journal}{\emph{ACL}}.
\newblock


\bibitem[\protect\citeauthoryear{Fiorini and Lu}{Fiorini and Lu}{2018}]%
        {Fiorini:18}
\bibfield{author}{\bibinfo{person}{Nicolas Fiorini} {and}
  \bibinfo{person}{Zhiyong Lu}.} \bibinfo{year}{2018}\natexlab{}.
\newblock \showarticletitle{Personalized neural language models for real-world
  query auto completion}. In \bibinfo{booktitle}{\emph{NAACL}}.
\newblock


\bibitem[\protect\citeauthoryear{Glorot and Bengio}{Glorot and Bengio}{2010}]%
        {glorot2010understanding}
\bibfield{author}{\bibinfo{person}{Xavier Glorot} {and} \bibinfo{person}{Yoshua
  Bengio}.} \bibinfo{year}{2010}\natexlab{}.
\newblock \showarticletitle{Understanding the difficulty of training deep
  feedforward neural networks}. In \bibinfo{booktitle}{\emph{AISTATS}}.
\newblock


\bibitem[\protect\citeauthoryear{Gutmann and Hyv{\"a}rinen}{Gutmann and
  Hyv{\"a}rinen}{2010}]%
        {gutmann2010}
\bibfield{author}{\bibinfo{person}{Michael Gutmann} {and} \bibinfo{person}{Aapo
  Hyv{\"a}rinen}.} \bibinfo{year}{2010}\natexlab{}.
\newblock \showarticletitle{Noise-contrastive estimation: A new estimation
  principle for unnormalized statistical models}. In
  \bibinfo{booktitle}{\emph{AISTATS}}.
\newblock


\bibitem[\protect\citeauthoryear{Hochreiter and Schmidhuber}{Hochreiter and
  Schmidhuber}{1997}]%
        {Hochreiter:97}
\bibfield{author}{\bibinfo{person}{Sepp Hochreiter} {and}
  \bibinfo{person}{Jürgen Schmidhuber}.} \bibinfo{year}{1997}\natexlab{}.
\newblock \showarticletitle{Long short-term memory}.
\newblock \bibinfo{journal}{\emph{Neural computation}} (\bibinfo{year}{1997}).
\newblock


\bibitem[\protect\citeauthoryear{Hofmann, Mitra, Radlinski, and
  Shokouhi}{Hofmann et~al\mbox{.}}{2014}]%
        {Hofmann:14}
\bibfield{author}{\bibinfo{person}{Kajta Hofmann}, \bibinfo{person}{Bhaskar
  Mitra}, \bibinfo{person}{Filip Radlinski}, {and} \bibinfo{person}{Milad
  Shokouhi}.} \bibinfo{year}{2014}\natexlab{}.
\newblock \showarticletitle{An eye-tracking study of user interactions with
  query auto completion}. In \bibinfo{booktitle}{\emph{CIKM}}.
\newblock


\bibitem[\protect\citeauthoryear{Jaech and Ostendorf}{Jaech and
  Ostendorf}{2018}]%
        {Jaech:18}
\bibfield{author}{\bibinfo{person}{Aaron Jaech} {and} \bibinfo{person}{Mari
  Ostendorf}.} \bibinfo{year}{2018}\natexlab{}.
\newblock \showarticletitle{Personalized Language Model for Query
  Auto-Completion}. In \bibinfo{booktitle}{\emph{ACL}}.
\newblock


\bibitem[\protect\citeauthoryear{Jiang, Chen, Cai, and Chen}{Jiang
  et~al\mbox{.}}{2018}]%
        {Jiang:18}
\bibfield{author}{\bibinfo{person}{Danyang Jiang}, \bibinfo{person}{Wanyu
  Chen}, \bibinfo{person}{Fei Cai}, {and} \bibinfo{person}{Honghui Chen}.}
  \bibinfo{year}{2018}\natexlab{}.
\newblock \showarticletitle{Neural Attentive Personalization Model for Query
  Auto-Completion}. In \bibinfo{booktitle}{\emph{IAEAC}}.
\newblock


\bibitem[\protect\citeauthoryear{Jiang, Ke, Chien, and Cheng}{Jiang
  et~al\mbox{.}}{2014}]%
        {Jiang:14}
\bibfield{author}{\bibinfo{person}{Jyun-Yu Jiang}, \bibinfo{person}{Yen-Yu Ke},
  \bibinfo{person}{Pao-Yu Chien}, {and} \bibinfo{person}{Pu-Jen Cheng}.}
  \bibinfo{year}{2014}\natexlab{}.
\newblock \showarticletitle{Learning user reformulation behavior for query
  auto-completion}. In \bibinfo{booktitle}{\emph{SIGIR}}.
\newblock


\bibitem[\protect\citeauthoryear{Kim, Jernite, Sontag, and Rush}{Kim
  et~al\mbox{.}}{2016}]%
        {Kim:16}
\bibfield{author}{\bibinfo{person}{Yoon Kim}, \bibinfo{person}{Yacine Jernite},
  \bibinfo{person}{David Sontag}, {and} \bibinfo{person}{Alexander~M Rush}.}
  \bibinfo{year}{2016}\natexlab{}.
\newblock \showarticletitle{Character-aware neural language models}. In
  \bibinfo{booktitle}{\emph{AAAI}}.
\newblock


\bibitem[\protect\citeauthoryear{LeCun and Bengio}{LeCun and Bengio}{1995}]%
        {LeCun:95}
\bibfield{author}{\bibinfo{person}{Yann LeCun} {and} \bibinfo{person}{Yoshua
  Bengio}.} \bibinfo{year}{1995}\natexlab{}.
\newblock \showarticletitle{Convolutional networks for images, speech, and time
  series}.
\newblock \bibinfo{journal}{\emph{The handbook of brain theory and neural
  networks}} (\bibinfo{year}{1995}).
\newblock


\bibitem[\protect\citeauthoryear{Li, Deng, Dong, Chang, Baeza-Yates, and
  Zha}{Li et~al\mbox{.}}{2017}]%
        {Li:17}
\bibfield{author}{\bibinfo{person}{Liangda Li}, \bibinfo{person}{Hongbo Deng},
  \bibinfo{person}{Anlei Dong}, \bibinfo{person}{Yi Chang},
  \bibinfo{person}{Ricardo Baeza-Yates}, {and} \bibinfo{person}{Hongyuan Zha}.}
  \bibinfo{year}{2017}\natexlab{}.
\newblock \showarticletitle{Exploring Query Auto-Completion and Click Logs for
  Contextual-Aware Web Search and Query Suggestion}. In
  \bibinfo{booktitle}{\emph{WWW}}.
\newblock


\bibitem[\protect\citeauthoryear{Li, Dong, Wang, Deng, Chang, and Zhai}{Li
  et~al\mbox{.}}{2014}]%
        {Li:14}
\bibfield{author}{\bibinfo{person}{Yanen Li}, \bibinfo{person}{Anlei Dong},
  \bibinfo{person}{Hongning Wang}, \bibinfo{person}{Hongbo Deng},
  \bibinfo{person}{Yi Chang}, {and} \bibinfo{person}{ChengXiang Zhai}.}
  \bibinfo{year}{2014}\natexlab{}.
\newblock \showarticletitle{A two-dimensional click model for query
  auto-completion}. In \bibinfo{booktitle}{\emph{SIGIR}}.
\newblock


\bibitem[\protect\citeauthoryear{Loshchilov and Hutter}{Loshchilov and
  Hutter}{2019}]%
        {Loshchilov2019DecoupledWD}
\bibfield{author}{\bibinfo{person}{Ilya Loshchilov} {and}
  \bibinfo{person}{Frank Hutter}.} \bibinfo{year}{2019}\natexlab{}.
\newblock \showarticletitle{Decoupled Weight Decay Regularization}. In
  \bibinfo{booktitle}{\emph{ICLR}}.
\newblock


\bibitem[\protect\citeauthoryear{Lowerre}{Lowerre}{1976}]%
        {Lowerre1976TheHS}
\bibfield{author}{\bibinfo{person}{Bruce~T. Lowerre}.}
  \bibinfo{year}{1976}\natexlab{}.
\newblock \showarticletitle{The HARPY speech recognition system}.
\newblock


\bibitem[\protect\citeauthoryear{Maxwell, Bailey, and Hawking}{Maxwell
  et~al\mbox{.}}{2017}]%
        {maxwell:2017}
\bibfield{author}{\bibinfo{person}{David Maxwell}, \bibinfo{person}{Peter
  Bailey}, {and} \bibinfo{person}{David Hawking}.}
  \bibinfo{year}{2017}\natexlab{}.
\newblock \showarticletitle{Large-scale generative query autocompletion}. In
  \bibinfo{booktitle}{\emph{ADCS}}.
\newblock


\bibitem[\protect\citeauthoryear{Mikolov, Karafi{\'a}t, Burget,
  {\v{C}}ernock{\`y}, and Khudanpur}{Mikolov et~al\mbox{.}}{2010}]%
        {mikolov2010}
\bibfield{author}{\bibinfo{person}{Tom{\'a}{\v{s}} Mikolov},
  \bibinfo{person}{Martin Karafi{\'a}t}, \bibinfo{person}{Luk{\'a}{\v{s}}
  Burget}, \bibinfo{person}{Jan {\v{C}}ernock{\`y}}, {and}
  \bibinfo{person}{Sanjeev Khudanpur}.} \bibinfo{year}{2010}\natexlab{}.
\newblock \showarticletitle{Recurrent neural network based language model}. In
  \bibinfo{booktitle}{\emph{InterSpeech}}.
\newblock


\bibitem[\protect\citeauthoryear{Mitra and Craswell}{Mitra and
  Craswell}{2015}]%
        {Mitra:15}
\bibfield{author}{\bibinfo{person}{Bhaskar Mitra} {and} \bibinfo{person}{Nick
  Craswell}.} \bibinfo{year}{2015}\natexlab{}.
\newblock \showarticletitle{Query auto-completion for rare prefixes}. In
  \bibinfo{booktitle}{\emph{CIKM}}.
\newblock


\bibitem[\protect\citeauthoryear{Mitra, Shokouhi, Radlinski, and Hofmann}{Mitra
  et~al\mbox{.}}{2014}]%
        {Mitra:14}
\bibfield{author}{\bibinfo{person}{Bhaskar Mitra}, \bibinfo{person}{Milad
  Shokouhi}, \bibinfo{person}{Filip Radlinski}, {and} \bibinfo{person}{Katja
  Hofmann}.} \bibinfo{year}{2014}\natexlab{}.
\newblock \showarticletitle{On user interactions with query auto-completion}.
  In \bibinfo{booktitle}{\emph{SIGIR}}.
\newblock


\bibitem[\protect\citeauthoryear{Mnih and Kavukcuoglu}{Mnih and
  Kavukcuoglu}{2013}]%
        {mnih2013}
\bibfield{author}{\bibinfo{person}{Andriy Mnih} {and} \bibinfo{person}{Koray
  Kavukcuoglu}.} \bibinfo{year}{2013}\natexlab{}.
\newblock \showarticletitle{Learning word embeddings efficiently with
  noise-contrastive estimation}. In \bibinfo{booktitle}{\emph{NIPS}}.
\newblock


\bibitem[\protect\citeauthoryear{Mnih and Teh}{Mnih and Teh}{2012}]%
        {mnih2012fast}
\bibfield{author}{\bibinfo{person}{Andriy Mnih} {and} \bibinfo{person}{Yee~Whye
  Teh}.} \bibinfo{year}{2012}\natexlab{}.
\newblock \showarticletitle{A fast and simple algorithm for training neural
  probabilistic language models}.
\newblock \bibinfo{journal}{\emph{arXiv preprint arXiv:1206.6426}}
  (\bibinfo{year}{2012}).
\newblock


\bibitem[\protect\citeauthoryear{Park and Chiba}{Park and Chiba}{2017}]%
        {Park:17}
\bibfield{author}{\bibinfo{person}{Dae~Hoon Park} {and} \bibinfo{person}{Rikio
  Chiba}.} \bibinfo{year}{2017}\natexlab{}.
\newblock \showarticletitle{A neural language model for query auto-completion}.
  In \bibinfo{booktitle}{\emph{SIGIR}}.
\newblock


\bibitem[\protect\citeauthoryear{Pass, Chowdhury, and Torgeson}{Pass
  et~al\mbox{.}}{2006}]%
        {pass2006picture}
\bibfield{author}{\bibinfo{person}{Greg Pass}, \bibinfo{person}{Abdur
  Chowdhury}, {and} \bibinfo{person}{Cayley Torgeson}.}
  \bibinfo{year}{2006}\natexlab{}.
\newblock \showarticletitle{A picture of search.}. In
  \bibinfo{booktitle}{\emph{InfoScale}}, Vol.~\bibinfo{volume}{152}.
\newblock


\bibitem[\protect\citeauthoryear{Sethy, Chen, Arisoy, and Ramabhadran}{Sethy
  et~al\mbox{.}}{[n. d.]}]%
        {sethy2015}
\bibfield{author}{\bibinfo{person}{Abhinav Sethy}, \bibinfo{person}{Stanley
  Chen}, \bibinfo{person}{Ebru Arisoy}, {and} \bibinfo{person}{Bhuvana
  Ramabhadran}.} \bibinfo{year}{[n. d.]}\natexlab{}.
\newblock \showarticletitle{Unnormalized exponential and neural network
  language models}. In \bibinfo{booktitle}{\emph{ICASSP}}.
\newblock


\bibitem[\protect\citeauthoryear{Shen, He, Gao, Deng, and Mesnil}{Shen
  et~al\mbox{.}}{2014}]%
        {Shen:14}
\bibfield{author}{\bibinfo{person}{Yelong Shen}, \bibinfo{person}{Xiaodong He},
  \bibinfo{person}{Jianfeng Gao}, \bibinfo{person}{Li Deng}, {and}
  \bibinfo{person}{Gr{\'e}goire Mesnil}.} \bibinfo{year}{2014}\natexlab{}.
\newblock \showarticletitle{Learning semantic representations using
  convolutional neural networks for web search}. In
  \bibinfo{booktitle}{\emph{WWW}}.
\newblock


\bibitem[\protect\citeauthoryear{Shokouhi}{Shokouhi}{2013}]%
        {Shokouhi:13}
\bibfield{author}{\bibinfo{person}{Milad Shokouhi}.}
  \bibinfo{year}{2013}\natexlab{}.
\newblock \showarticletitle{Learning to personalize query auto-completion}. In
  \bibinfo{booktitle}{\emph{SIGIR}}.
\newblock


\bibitem[\protect\citeauthoryear{Shokouhi and Radinsky}{Shokouhi and
  Radinsky}{2012}]%
        {Shokouhi:12}
\bibfield{author}{\bibinfo{person}{Milad Shokouhi} {and} \bibinfo{person}{Kira
  Radinsky}.} \bibinfo{year}{2012}\natexlab{}.
\newblock \showarticletitle{Time-sensitive query auto-completion}. In
  \bibinfo{booktitle}{\emph{SIGIR}}.
\newblock


\bibitem[\protect\citeauthoryear{Sundermeyer, Schl{\"u}ter, and
  Ney}{Sundermeyer et~al\mbox{.}}{2012}]%
        {sundermeyer2012}
\bibfield{author}{\bibinfo{person}{Martin Sundermeyer}, \bibinfo{person}{Ralf
  Schl{\"u}ter}, {and} \bibinfo{person}{Hermann Ney}.}
  \bibinfo{year}{2012}\natexlab{}.
\newblock \showarticletitle{LSTM neural networks for language modeling}. In
  \bibinfo{booktitle}{\emph{InterSpeech}}.
\newblock


\bibitem[\protect\citeauthoryear{Vaswani, Shazeer, Parmar, Uszkoreit, Jones,
  Gomez, Kaiser, and Polosukhin}{Vaswani et~al\mbox{.}}{2017}]%
        {vaswani2017attention}
\bibfield{author}{\bibinfo{person}{Ashish Vaswani}, \bibinfo{person}{Noam
  Shazeer}, \bibinfo{person}{Niki Parmar}, \bibinfo{person}{Jakob Uszkoreit},
  \bibinfo{person}{Llion Jones}, \bibinfo{person}{Aidan~N Gomez},
  \bibinfo{person}{{\L}ukasz Kaiser}, {and} \bibinfo{person}{Illia
  Polosukhin}.} \bibinfo{year}{2017}\natexlab{}.
\newblock \showarticletitle{Attention is all you need}. In
  \bibinfo{booktitle}{\emph{NIPS}}.
\newblock


\bibitem[\protect\citeauthoryear{Vaswani, Zhao, Fossum, and Chiang}{Vaswani
  et~al\mbox{.}}{2013}]%
        {vaswani2013}
\bibfield{author}{\bibinfo{person}{Ashish Vaswani}, \bibinfo{person}{Yinggong
  Zhao}, \bibinfo{person}{Victoria Fossum}, {and} \bibinfo{person}{David
  Chiang}.} \bibinfo{year}{2013}\natexlab{}.
\newblock \showarticletitle{Decoding with large-scale neural language models
  improves translation}. In \bibinfo{booktitle}{\emph{EMNLP}}.
\newblock


\bibitem[\protect\citeauthoryear{Wang, Zhang, Mohan, Dhillon, and Kolter}{Wang
  et~al\mbox{.}}{2018}]%
        {Wang2018}
\bibfield{author}{\bibinfo{person}{Po-Wei Wang}, \bibinfo{person}{Huan Zhang},
  \bibinfo{person}{Vijai Mohan}, \bibinfo{person}{Inderjit~S. Dhillon}, {and}
  \bibinfo{person}{J.~Zico Kolter}.} \bibinfo{year}{2018}\natexlab{}.
\newblock \showarticletitle{Realtime query completion via deep language
  models}. In \bibinfo{booktitle}{\emph{SIGIR eCom}}.
\newblock


\bibitem[\protect\citeauthoryear{Whiting and Jose}{Whiting and Jose}{2014}]%
        {Whiting:14}
\bibfield{author}{\bibinfo{person}{Stewart Whiting} {and}
  \bibinfo{person}{Joemon~M Jose}.} \bibinfo{year}{2014}\natexlab{}.
\newblock \showarticletitle{Recent and robust query auto-completion}. In
  \bibinfo{booktitle}{\emph{WWW}}.
\newblock


\bibitem[\protect\citeauthoryear{Williams and Zipser}{Williams and
  Zipser}{1989}]%
        {Williams:89}
\bibfield{author}{\bibinfo{person}{Ronald~J Williams} {and}
  \bibinfo{person}{David Zipser}.} \bibinfo{year}{1989}\natexlab{}.
\newblock \showarticletitle{A learning algorithm for continually running fully
  recurrent neural networks}.
\newblock \bibinfo{journal}{\emph{Neural computation}} \bibinfo{volume}{1},
  \bibinfo{number}{2} (\bibinfo{year}{1989}).
\newblock


\bibitem[\protect\citeauthoryear{Zhang, Goyal, Kong, Deng, Dong, Chang, Gunter,
  and Han}{Zhang et~al\mbox{.}}{2015}]%
        {Zhang:15}
\bibfield{author}{\bibinfo{person}{Aston Zhang}, \bibinfo{person}{Amit Goyal},
  \bibinfo{person}{Weize Kong}, \bibinfo{person}{Hongbo Deng},
  \bibinfo{person}{Anlei Dong}, \bibinfo{person}{Yi Chang},
  \bibinfo{person}{Carl~A Gunter}, {and} \bibinfo{person}{Jiawei Han}.}
  \bibinfo{year}{2015}\natexlab{}.
\newblock \showarticletitle{adaqac: Adaptive query auto-completion via implicit
  negative feedback}. In \bibinfo{booktitle}{\emph{SIGIR}}.
\newblock


\end{thebibliography}
\end{document}